\documentclass[12pt]{article}
\usepackage{amsmath,amssymb,amsthm,amsfonts}
\usepackage{arxiv}
\usepackage{hyperref}
\usepackage{nicefrac}
\usepackage{bbm}
\usepackage{mathrsfs}
\usepackage{algorithm}
\usepackage{algorithmic}
\usepackage{times}
\usepackage{graphicx}
\usepackage{subcaption}
\usepackage{url}
\usepackage{booktabs}
\usepackage{secdot}
\usepackage{multirow}
\usepackage{multicol}

\usepackage{hyperref}
\usepackage{url}
\usepackage{xfrac,nicefrac}
\usepackage{algorithm}
\usepackage{algorithmic}
\usepackage{multirow}
\usepackage{graphicx}
\usepackage{booktabs}
\usepackage{dsfont}
\usepackage{subcaption}

\newtheorem{theorem}{Theorem}
\newtheorem{lemma}{Lemma}

\usepackage{amsmath,amsfonts,bm}

\def\eqref#1{equation~\ref{#1}}

\def\1{\bm{1}}

\DeclareMathAlphabet{\mathsfit}{\encodingdefault}{\sfdefault}{m}{sl}
\SetMathAlphabet{\mathsfit}{bold}{\encodingdefault}{\sfdefault}{bx}{n}

\newcommand{\R}{\mathbb{R}}

\title{Dynamic Low-rank Approximation of Full-Matrix Preconditioner for Training Generalized Linear Models}
\usepackage{times}

\author{
Tatyana Matveeva\thanks{Corresponding author} \\
Skoltech, Moscow, Russia\\
HSE, Moscow, Russia\\
\url{Tatyana.Matveeva@skoltech.ru}
\And
Aleksandr Katrutsa\\
Avito, Moscow, Russia\\
Skoltech, Moscow, Russia\\
\url{amkatrutsa@gmail.com} 
\And
Evgeny Frolov\\
AIRI, Moscow, Russia\\
Skoltech, Moscow, Russia\\
\url{frolov@airi.net}
}

\begin{document}

\maketitle

\begin{abstract}
Adaptive gradient methods like Adagrad and its variants are widespread in large-scale optimization. 
However, their use of diagonal preconditioning matrices limits the ability to capture parameter correlations. 
Full-matrix adaptive methods, approximating the exact Hessian, can model these correlations and may enable faster convergence. 
At the same time, their computational and memory costs are often prohibitive for large-scale models. 
To address this limitation, we propose AdaGram, an optimizer that enables efficient full-matrix adaptive gradient updates. 
To reduce memory and computational overhead, we utilize fast symmetric factorization for computing the preconditioned update direction at each iteration. 
Additionally, we maintain the low-rank structure of a preconditioner along the optimization trajectory using matrix integrator methods. 
Numerical experiments on standard machine learning tasks show that AdaGram converges faster or matches the performance of diagonal adaptive optimizers when using rank five and smaller rank approximations. 
This demonstrates AdaGram’s potential as a scalable solution for adaptive optimization in large models.
\end{abstract}

\section{Introduction}
The landscape of stochastic optimization has been fundamentally transformed by the introduction of adaptive gradient methods, which automatically adjust learning rates based on the geometry of the optimization problem. 
In classical gradient-based optimization, the selection of a learning rate~$\eta$ is critical; an improper choice can lead to slow convergence or divergence of the optimization trajectory. 
A common desideratum is for the step size to decrease as the optimizer approaches a minimum. 
To enforce this, various learning rate decay schedulers are employed. 
These schedulers systematically reduce the learning rate $\eta$ over time. 
Common strategies include step decay, where the learning rate is reduced by a fixed factor at specific epochs; exponential decay, which applies a continuous reduction of the form $\eta_t = \eta_0 \exp(-kt)$; and cosine annealing, which follows a sinusoidal trajectory from the initial rate to a minimum value. 
While effective, these schedulers are typically non-adaptive, as their evolution is pre-determined and does not respond to the observed gradient statistics during training.

A parallel line of inquiry for improving upon vanilla SGD is the incorporation of momentum. 
Rather than relying solely on the current gradient $g_t$ at step $t$, momentum methods accumulate an exponentially decaying moving average of past gradients, often termed velocity $v_t$. 
The parameter update is then a function of this velocity, i.e., $\theta_{t+1} = \theta_t - \eta v_t$. 
This has the effect of dampening high-frequency oscillations along directions of high curvature while accelerating progress along directions with persistent, low-variance gradients. 
This interpretation becomes especially evident if momentum-based methods are analyzed from the ODE discretization perspective~\cite{su2016differential,laborde2020lyapunov,leplat2022nag}.
The canonical example is SGD with Momentum, with a notable refinement being the Nesterov Accelerated Gradient (NAG)~\cite{nesterov1983method}, which calculates the gradient after applying a preliminary momentum step, providing a more effective "lookahead" mechanism.

Adaptive methods, as introduced, directly address the limitation of a single, static learning rate by maintaining per-parameter learning rates scaled by historical gradient information. These methods address a critical limitation of traditional stochastic gradient descent (SGD): the use of a single learning rate for all parameters, regardless of their individual optimization characteristics. The pioneering method, AdaGrad, introduced in~\cite{duchi2011adagrad}, accumulates the sum of squared gradients for each parameter and normalizes the update by the square root of this sum. While highly effective for sparse problems, its monotonically decreasing learning rate can prematurely halt training. RMSProp refines this by using an exponentially decaying moving average of squared gradients, preventing the aggressive decay. The most prominent adaptive method, Adam (Adaptive Moment Estimation)~\cite{kingma2015adam}, synergistically combines the momentum concept (a first-moment estimate) with the per-parameter scaling of RMSProp (a second-moment estimate~\cite{hinton2012rmsprop}), establishing itself as a default optimizer for a vast range of deep learning applications.

In this work, we exclude methods enriched with momentum and exponential decay techniques from our scope and develop a method based solely on Adagrad that considers feature correlations. 
The method, introduced in this work, can be further developed with techniques used in more sophisticated adaptive algorithms (Adam~\cite{kingma2015adam}, RMSProp~\cite{hinton2012rmsprop}, etc.)

Our contributions are:
\begin{itemize}
    \item AdaGram, an optimization algorithm that enables efficient full-matrix adaptive gradient updates. We use the fast symmetric factorization technique, introduced in~\cite{ambikasaran2014fast}, for the computation of the preconditioned update direction at each iteration. Additionally, we link the iterative nature of the preconditioner with matrix integrator theory and decomposition update techniques to avoid costly recomputation of the matrix decomposition at each step.
    \item An empirical validation of AdaGram's performance and efficiency. We conduct numerical experiments on Generalized Linear Models (GLMs), specifically a logistic regression task using a single-layer perceptron. The results, presented in Section~\ref{experiments}, demonstrate the practical advantages of our method and motivate its generalization to more complex, arbitrary model architectures.
\end{itemize}

\section{Related work}

Classical stochastic gradient descent (SGD) faces several fundamental limitations that impact its optimization performance. To address these, the full-matrix AdaGrad algorithm, introduced in~\cite{duchi2011adagrad} and represented in Algorithm~\ref{alg:adagrad-matrix}, marked a significant step forward. By capturing the complete second-order information from the gradient history, including correlations and cross-parameter dependencies, this approach enables sophisticated preconditioning that accounts for the geometry of the loss landscape. However, the same work also recognized the practical challenge of this method and proposed a diagonal preconditioning strategy as a computationally feasible alternative. In this variant, each parameter receives an individualized learning rate based only on its own gradient history. While efficient, this simplification comes at the cost of ignoring the rich information contained in parameter interactions.

The computational intractability of full-matrix methods in high-dimensional settings stems from their prohibitive complexity. These approaches necessitate matrix inversion and square root operations, resulting in an $O(n^3)$ computational cost per iteration. This cubic scaling renders such methods impractical for large-scale applications, where diagonal approximations with linear $O(n)$ complexity remain the standard.

Consequently, much research has focused on making full-matrix methods practical. Initial explorations sought to bypass the direct computation of the inverse square root through scalable approximations. For instance, the study~\cite{krummenacher2016scalableadaptivestochasticoptimization} proposed maintaining a low-rank factorization of the preconditioning matrix. While this avoids cubic complexity, it introduces challenges in maintaining an accurate factorization. Similarly, the study~\cite{jain2018neural} explored online matrix estimation to approximate the inverse Hessian directly but found that the high variance of stochastic gradients led to unstable estimates, degrading optimizer performance.

A significant conceptual advance came with the Shampoo algorithm~\cite{gupta2018shampoopreconditionedstochastictensor}, which introduced a novel preconditioning strategy based on Kronecker factorization. Instead of approximating the full covariance matrix, Shampoo leverages the tensor structure of gradients in deep learning by maintaining separate, smaller preconditioners for the input and output dimensions. The core innovation is applying a $-\frac{1}{4}$ power to each factor, such that their combined effect approximates the inverse square root of the full covariance matrix. This dramatically reduces computational overhead while capturing crucial geometric information. More recent variants like SOAP~\cite{vyas2024soap} have built upon this foundation, achieving near-linear time complexity and making them competitive with Adam in wall-clock time.

Other promising research directions have emerged concurrently. The paper~\cite{agarwal2020efficientfullmatrixadaptiveregularization} revisited the full-matrix challenge by combining efficient regularization with low-rank approximations. However, despite strong theoretical convergence guarantees, the algorithm did not outperform established benchmarks in practical wall-clock time. Another approach, proposed in~\cite{choudhury2024remove}, involves removing the square root from the update rule entirely. 
This modification not only improves empirical performance but also endows the resulting algorithm with scale-invariance, a highly desirable property for stabilizing training.

To address the aforementioned limitations, we propose an efficient algorithm for the iterative computation of the inverse square root of the gradient second-moment matrix. Our approach leverages recent advances in numerical linear algebra, exploiting the low-rank structure inherent in gradient accumulation over finite time windows. The method circumvents the computational bottleneck of direct matrix inversion through specialized techniques for fast matrix function evaluation, thereby making a full-matrix approach feasible.

The foundational principles explored in this work are broadly applicable. These ideas can be integrated with or seen as extensions to a wide range of established first-order adaptive methods~--- including Adam~\cite{kingma2015adam}, RMSprop~\cite{hinton2012rmsprop}, Adadelta~\cite{zeiler2012adadelta}, and Nadam~\cite{dozat2016nadam}---offering a rich avenue for future hybrid algorithms.



\section{Adagram: adaptive optimizer based on the low-rank integrator }
\label{headings}

This section presents the mathematical framework and algorithmic components for constructing an adaptive optimizer that enables efficient full-matrix adaptive gradient updates, leveraging the special structure of a preconditioner matrix.

\subsection{Preliminaries}

Following the seminal work~\cite{duchi2011adagrad}, we define the general schedule for adaptive gradient step updates as:
\begin{equation}
    \theta \leftarrow \theta - \eta\,G_t^{-\nicefrac{1}{2}}g_t,
\end{equation}
where $\theta \in \R^{n}$ is a parameter vector that we optimize with the preconditioned gradient descent, $g_t$ is a gradient at step~$t$, and $\eta$ is a scalar denoting the learning rate. 
An updatable preconditioner matrix $G_t$ has the form
\begin{equation}\label{eq:gram_matrix}
    G_t = \epsilon I + \sum_{\tau=1}^t g_\tau g_\tau^\top.
\end{equation}
It can be equivalently rewritten via the following rank-1 update series
\begin{equation}\label{eq:gram_matrix}
    G_{t+1} = G_t + g_{t+1}^{} g_{t+1}^\top,
\end{equation}
initialized by $G_0 = \epsilon I$, where $\epsilon$ is a scalar and $I$ is an $n\times n$ identity matrix.

We denote an ordinary $L_2$ vector norm as $\|\cdot\|$ and use $\text{vec()}$ notation to denote the flattening a matrix (a tensor in general case) into a column vector. 
By construction, the existence of an inverse of matrix $G_t$ is guaranteed by the addition of an $\epsilon$ to the matrix main diagonal on the step $0$. 
The square root exists as the matrix $G_t$ is positive semi-definite by design.

Commonly, only the diagonal part of $G_t$ is considered during optimization to ensure fast parameter updates at each step. 
On the other hand, the full-matrix adaptive optimization can potentially bring faster convergence by incorporating the richer second-order information into the convergence process. 
However, direct implementation of such an approach incurs a prohibitive $O(n^3)$ computational cost of inverting the full matrix with additional $O(n^2)$ memory overhead for storing it.

To address the aforementioned computational challenges, we propose a \emph{novel algorithm that utilizes the incremental nature of the matrix $G_t$ along with the special dynamical low-rank representation for efficiently updating the preconditioner}. 
The method maintains the theoretical advantages of full-matrix preconditioning while achieving better computational efficiency. 
It \emph{enables additional control over the balance between the second-order accuracy and practical scalability in large-scale optimization}. 
To achieve this, we revisit the problem under the isometrical equivalence condition, which helps lift strict constraints that prevent the derivation of an efficient computational form of the preconditioner. 

\subsection{Revisiting the square root computation}
Consider the task of incremental computation of the square root of the matrix $G_t$ at two consecutive time steps. 
Having computed the matrix square root $G_t^{\nicefrac{1}{2}}$ at time $t$, one can derive the following form for the factors of $G_{t+1}$ at time $t+1$:
\begin{equation}\label{eq:symroot}
    G_t + g_{t+1} g^\top_{t+1} = G^{\nicefrac{1}{2}}_{t}\left(I + \bar{g}_{t+1} \bar{g}^{\top}_{t+1} \right)G^{\nicefrac{1}{2}}_{t} 
    = G^{\nicefrac{1}{2}}_{t}\left(I + \alpha \bar{g}_{t + 1} \bar{g}^{\top}_{t + 1} \right)\left(I + \alpha \bar{g}_{t+1} \bar{g}^{\top}_{t + 1} \right)G^{\nicefrac{1}{2}}_{t},
\end{equation}
where $\bar{g}_{t+1} = G_t^{-1}g_{t+1}$ $\alpha$ is a scalar constant that allows factorizing the term $\left(I + \bar{g} \bar{g}^{\top} \right)$ into the product of two matrices, it can be found from the identity:
\begin{equation}\label{eq:rootconst}
    1 + \alpha \| \bar{g} \|^2 = \left( 1 + \| \bar{g} \|^2 \right)^{\nicefrac{1}{2}}.
\end{equation}
From Eq.~\ref{eq:symroot} it follows that even if we knew the exact symmetric matrix square root of $G_t$ at some moment $t$, this knowledge could not be directly reused for obtaining the matrix square root at the next step as it would require matrices $G_t^{\nicefrac{1}{2}}$ and $(I + \bar{g} \bar{g}^\top)$ to commute, which does not hold in the general case, i.e.:
\begin{equation}
    G_t^{\nicefrac{1}{2}}(I + \alpha \bar{g}_{t+1} \bar{g}_{t+1}^{\top}) \neq (I + \alpha \bar{g}_{t+1} \bar{g}_{t+1}^{\top})G^{\nicefrac{1}{2}} \neq (G_t + g_{t+1} g^\top_{t+1})^{\nicefrac{1}{2}}.
\end{equation}
However, it also hints that computing the non-symmetric triangular factors $L_t$ of the Cholesky decomposition $G = L_t L_t^\top$ instead of symmetric $G^{\nicefrac{1}{2}}$ offers a viable solution. 
Although Cholesky factors exhibit different properties compared to symmetric matrix square roots, the substitution still maintains equivalence under the isometry condition, which is directly verified by the expression $\| G^{-\nicefrac{1}{2}} g \|^2 = g^\top G^{-1} g = g^\top (L L^\top)^{-1} g = \| L^{-1} g \|^2$.

Conveniently, as we will demonstrate below, both the Cholesky factor $L_t$ and its inverse $L_t^{-1}$ can be obtained analytically for any $t$ in the form of a matrix factorization with special structure. 
We will show how to improve the scalability of this new representation further using a dynamical low-rank approximation scheme.

\subsection{Efficient Cholesky Factor Computation}
Effective optimization requires computation of the preconditioned gradient, which involves a matrix-vector product. We demonstrate that this preconditioned vector can be computed recursively. Furthermore, through matrix decomposition techniques, computational complexity can be substantially reduced by operating exclusively on low-rank matrices and vectors.

\begin{theorem}[Recursive computation of a preconditioned gradient]
\label{thm:recursive_grad}

Let $g_{t +1}$ be the gradient on step $t+1$ and $L_t$ be the Cholesky factor of a matrix $G$ on step $t$. Then the preconditioned gradient on step $t+1$ can be calculated as follows:
\begin{equation}\label{eq:matvec_recusrion}
    L_{t+1}^{-1}g_{t+1} = \frac{1}{\sqrt{1 + \|L_t^{-1}g_{t+1}\|^2}}L_t^{-1}g_{t+1}.
\end{equation}
\end{theorem}

\begin{proof}[Proof Sketch]
Following the symmetric factorization approach similarly to Eq.~\ref{eq:symroot}, one gets an explicit recursive expression for the Cholesky factor of $G_{t+1}$:
\begin{equation}\label{eq:root_recursive}
    L_{t+1} = L_t\left(I+\alpha_{t+1}\bar{g}_{t+1}\bar{g}_{t+1}^\top\right),
\end{equation}
where $\bar{g}_{t+1}=L_t^{-1}g_{t+1}$ and $\alpha_{t+1}$ satisfies Eq.~\ref{eq:rootconst}.

Applying the Sherman-Woodberry-Morrison identity to Eq.~\ref{eq:root_recursive} yields:
\begin{equation}\label{eq:root}
    L_{t+1}^{-1} = \left(I - \beta_{t+1}^{}\bar{g}_{t+1}^{}\bar{g}_{t+1}^\top\right)L_{t}^{-1},
\end{equation}
where $\beta_{t+1} = \frac{\alpha_{t+1}}{1+\alpha_{t+1}\|\bar{g}_{t+1}\|^2}$. The full proof is presented in Appendix~\ref{theorem_proof}.
\end{proof}

The scheme provided above significantly simplifies the preconditioning process. 
However, in practice, keeping the matrix $L_t$ in memory requires $O(n^2)$ space; moreover,  multiplication by a vector still results in $2n^2 + n$, leading to undesirable computational costs. 
This estimation can be improved by introducing low-rank factors, which we will update ``on-the-fly'' with the projector splitting algorithm.

Unraveling the recursive definition of Eq.~\ref{eq:root} gives rise to the following lemma.

\begin{lemma}[Recursive representation of the preconditioner inverse]\label{lem:lemma1}

Let $P_t$ and $Q_t$ be iteratively built matrices of the form:
\begin{align*}
    P_{t+1} &= \left[P_t\quad\beta_{t+1}{\bar g}_{t+1}\right], \\
    Q_{t+1} &= \left[Q_t\quad\left(I-Q_t^{}P_t^\top\right){\bar g}_{t+1}\right].
\end{align*}
Then the preconditioner matrix $L^{-1}$ on step $t+1$ takes the form:
\begin{equation}\label{eq:propagated}
    L_{t+1}^{-1} = \left(I-P^{}_{t+1}Q_{t+1}^\top\right)L_0^{-1}.
\end{equation}
\begin{proof}
    The proof is presented in Appendix~\ref{lemma_proof}.
\end{proof}
\end{lemma}

Note that matrices $P_t$ and $Q_t$ grow with time, thus requiring an increasing amount of memory to support recursive computations.

\subsection{Updating preconditioner with low-rank integrator}

Following Lemma~\ref{lem:lemma1}, we obtain a recursive procedure for computing the inverse of the Cholesky factor using the matrices $P_t$ and $Q_t$. The primary complication of this approach is the unbounded growth in the dimensionality of these factors over time.

One way to restrict the growth is to perform a low-rank approximation of the $P_t Q_t^\top$ term via SVD. 
There are schemes that improve the efficiency of this step by utilizing the special low-rank structure of the updates that allow avoiding recomputation of the full SVD at each step, for example, the algorithm given in~\cite{Brand2006}.
Another way is to perform updates using a dynamic low-rank approximation, via \emph{matrix integrator approach} introduced in~\cite{lubich2013projectorsplittingintegratordynamicallowrank}.  


Let us have a closer look on the Lemma~\ref{lem:lemma1} base step:

\begin{align*}
L_{t+1}^{-1} &= (I - P_t Q_t^T) L_0^{-1} \left(I - \beta_{t+1} \bar{g}_{t+1} \bar{g}_{t+1}^T\right)\\
&= (I - P_t Q_t^T - \beta_{t+1} \bar{g}_{t+1} \bar{g}_{t+1}^T + P_t Q_t^T \beta_{t+1} \bar{g}_{t+1} \bar{g}_{t+1}^T) L_0^{-1}\\
 &= (I - \underbrace{(P_tQ_t^\top + \beta_{t+1}^{}\bar{g}_{t+1}^{}\bar{g}_{t+1}^\top\left(I-P_{t}^{}Q_{t}^\top\right))}_{P_{t+1}Q_{t+1}^\top}).
\end{align*}

Let 
\begin{equation}
    A_{t} = P_{t}^{}Q_{t}^\top
\end{equation} 
and 
\begin{equation}\label{eq:update}
    \Delta A_{t+1} =\beta_{t+1}^{}\bar{g}_{t+1}^{}\bar{g}_{t+1}^\top\left(I-P_{t}^{}Q_{t}^\top\right).
\end{equation}

Then, 

\begin{equation}\label{eq:psi}
P_{t+1}Q_{t+1}^\top = 
    A_{t+1} = A_t + \Delta A_{t+1}.
\end{equation}

The resulting form can be interpreted as a matrix integration process. The method allows for modeling matrix dynamics over steps by recalculating the rank-r SVD decomposition
$P_0Q_0^\top = U_0 S_0 V_0^\top$ with respect to the update from Eq.~\ref{eq:update}.

For the implementation, we used the full practical algorithm from section 3.2 of the~\cite{lubich2013projectorsplittingintegratordynamicallowrank}, which is given in Algorithm~\ref{alg:first_order_split}.

\begin{algorithm}
\caption{Projector splitting method, practical algorithm}
\label{alg:first_order_split}
\begin{algorithmic}[1]
\STATE \textbf{Given:} A rank-\(r\) matrix factorization \(A_t = U_0 S_0 V_0^\top\) and the increment \(\Delta A\).
\STATE Set \(K_1 = U_0 S_0 + \Delta A V_0\).
\STATE Compute the factorization \(U_1 \tilde{S}_1 = K_1\) (e.g., using QR or SVD), where \(U_1\) has orthonormal columns and \(\tilde{S}_1\) is an \(r \times r\) matrix.
\STATE Set \(\tilde{S}_0 = \tilde{S}_1 - U_1^\top \Delta A V_0\).
\STATE Set \(L_1 = V_0 \tilde{S}_0^\top + \Delta A^\top U_1\).
\STATE Compute the factorization \(V_1 S_1^\top = L_1\) (e.g., using QR or SVD), with \(V_1\) having orthonormal columns and \(S_1\) being an \(r \times r\) matrix.
\STATE \textbf{Output:} The new rank-\(r\) approximation \(Y_1 = U_1 S_1 V_1^\top\).
\end{algorithmic}
\end{algorithm}

A key advantage of this approach lies in its adaptive update mechanism. The magnitude of decomposition modification is directly proportional to the magnitude of the gradient update. This property is particularly beneficial for optimization, as it enables the method to ``forget'' about the large initial updates that are typically irrelevant to later epochs.

Additionally, a hyperparameter $\mu$ can be incorporated in Eq.~\ref{eq:psi} to control the relative weighting between current gradient updates and historical information accumulated from previous epochs. 
The optimal value of $\mu$ should be determined before training based on the characteristics of the loss landscape.

\begin{equation}\label{eq:psi_moment}
P_{t+1}Q_{t+1}^\top = 
    A_{t+1} = \mu A_t + (1 - \mu)\Delta A_{t+1}.
\end{equation}
The resulting AdaGram optimizer is presented in Algorithm~\ref{alg:adagram-matrix}. 

\begin{algorithm}[!h]
\caption{AdaGram}
\label{alg:adagram-matrix}
\begin{algorithmic}[1]
\REQUIRE $\eta > 0$, $\epsilon > 0$, $\beta_t$ (learning rate for preconditioner)
\STATE Initialize $W_1 = \bm{0}_{m \times n}$; $L_0 = \sqrt{\epsilon} \bm{I}_{m}$; \COMMENT{$G_t = L_tL_t^\top \in \mathbb{R}^{m \times m}$}
\FOR{$t = 1$ \TO $T$}
    \STATE Receive loss function $f_t : \mathbb{R}^{m \times n} \to \mathbb{R}$
    \STATE Compute gradient $\bm{g}_t =  \text{vec}(\nabla f_t(W_t))$ \COMMENT{$\bm{g}_t \in \mathbb{R}^{mn}$}
    \STATE Compute transformed gradient: $\bar{\bm{g}}_t = L_{t-1}^{-1}\bm{g}_t = (I - P_{t-1}Q_{t-1}^\top)L_0^{-1}\bm{g}_t$
    \STATE Update $P$ and $Q$ \COMMENT{see Eq.~\ref{eq:psi_moment} and Algorithm~\ref{alg:first_order_split}}
    \STATE Update parameters: $\bm{W}_{t+1} = \text{unvec}(\bm{w}_t - \eta \bar{\bm{g}}_t/\sqrt{1 + \|\bar{\bm{g}}_t\|^2}$) \COMMENT{follows Eq.~\ref{eq:matvec_recusrion}}
\ENDFOR
\end{algorithmic}
\end{algorithm}

\section{Numerical experiments}
\label{experiments}
This section presents our experimental evaluation of the AdaGram algorithm in the following Section. 
All experiments were conducted on a cluster with Tesla V100-SXM3-32GB, Python 3.9, and PyTorch 2.5.1.

\subsection{Training generalized linear models}

For the consistent hypothesis confirmation, we consider only generalized linear models (GLM), specifically the logistic regression case. 
Below, we summarize some classical results for these models.  

A framework of generalized linear models introduced in~\cite{nelderwedderburn1972glm} generalizes traditional linear regression models by allowing the response variable to follow distributions from the exponential family. 
The linear combination of predictors is linked to the mean of the response through a monotonic link function.

In binary logistic regression, the target variable $y$ is modeled by a Bernoulli distribution with probability  $p = \sigma(\theta^\top x) = \frac{\exp(\theta^\top x)}{1 + \exp(\theta^\top x)}$. 
The Hessian matrix $\mathbf{H}$ of the cross-entropy loss function for the batch case looks as follows:





\begin{equation}
\label{eq:batch_hessian}
\mathbf{H} = \frac{1}{B}\mathbf{X}^T \mathbf{P} \mathbf{X}.
\end{equation}
This formula resembles the AdaGrad preconditioner matrix, given in Eq.~\ref{eq:gram_matrix} and motivates such preconditioner choice.

\subsection{Experimental Setup}
For empirical validation of the proposed algorithms, we adapt several UCI~\cite{UCI} datasets (Heart, Australian, and Splice) from the experimental framework of~\cite{choudhury2024remove}. 
Their main characteristics are shown in the Table~\ref{tab:dataset_stats}.

\begin{table}[!h]
\centering
\caption{Statistics of the UCI Datasets Used in This Work}
\label{tab:dataset_stats}
\begin{tabular}{lrrr}
\toprule
\textbf{Dataset} & \textbf{Instances} & \textbf{Features} & \textbf{Classes} \\
\midrule
Australian Credit & 690 & 14 & 2 \\
Heart Disease     & 303 & 13 & 2 \\
Splice Junctions  & 3,186 & 60 & 3 \\
\bottomrule
\end{tabular}
\end{table}

The baseline methods are the following: Vanilla SGD, Diagonal AdaGrad~\cite{duchi2011adagrad},
Shampoo~\cite{gupta2018shampoopreconditionedstochastictensor},
KATE~\cite{choudhury2024remove}. 
Hyperparameter selection for each dataset was performed via grid search over batch size, learning rate, and numerical stability parameter $\epsilon$. 
For AdaGram variants, additional grid search was conducted over rank $r$ and memory parameter $\mu$. 
The configuration that minimized the objective function value at the final training epoch was selected for all reported experiments. 
The source code for the experiments is available in the following GitHub repository~\url{https://github.com/tnmtvv/AdaGram}.

\subsection{Evaluation on synthetic datasets}
These experiments aim to establish key algorithmic properties through controlled evaluation. 
The incorporation of the full gradient covariance matrix is designed to capture inter-parameter correlations, and we seek to validate this capability for our proposed methods empirically.
In the case of logistic regression, gradients are linked to the samples as follows

\[
\nabla_{\theta} J_B(\theta) = \frac{1}{b} \sum_{i \in B} \left( \sigma(\theta^T x^{(i)}) - y^{(i)} \right) x^{(i)}.
\]

\begin{figure}[h]
    \centering
    \includegraphics[width=0.8\linewidth]{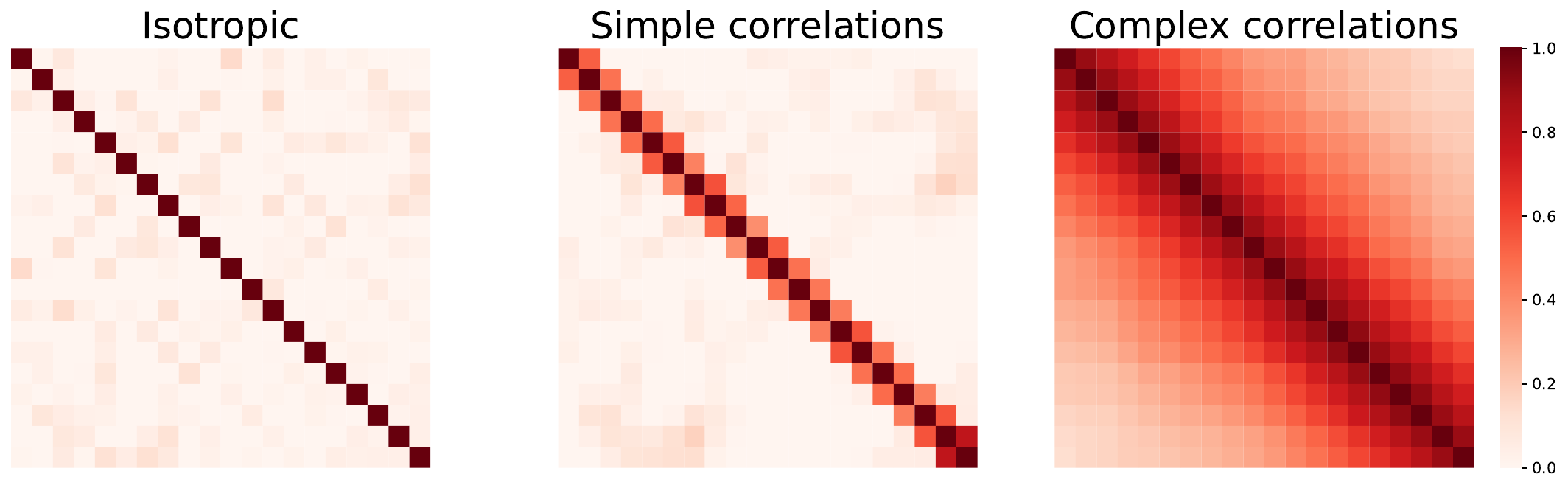}
    \caption{Correlation matrices}
    \label{fig:corr_matrices}
\end{figure}

Thus, we can control the complexity of the loss landscape by changing the feature correlations.
Following~\cite{katrutsa2015stress}, we introduce the following structures of the synthetic datasets and compare the optimizer's behavior for them:
\begin{itemize}
    \item Isotropic -- a dataset with independent features. 
    We expect simple methods, such as vanilla SGD and Adagrad, to be the fastest to converge in this scenario.
    \item Data with simple correlations -- dataset with a three-diagonal correlation matrix, which induces structured diagonal dominant correlations. 
    In this case, we expect to see the compatible or supreme convergence speed of the full-matrix approximating methods over the simple ones.
    \item Data with complex correlation -- we consider a borderline case with highly correlated features and expect to see the visible gain in full-matrix approximating algorithms' performance. 
\end{itemize}
The corresponding correlation matrices are visualized in Figure~\ref{fig:corr_matrices}.

\begin{figure}[h]
    \centering
    \includegraphics[width=\linewidth]{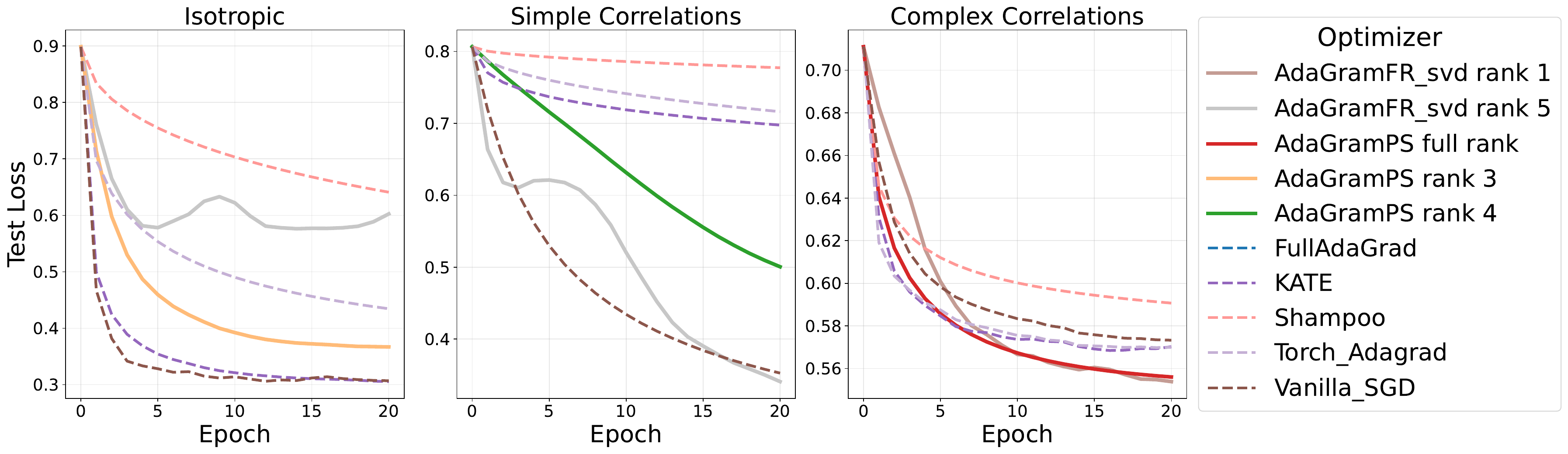}
    \caption{ The plots compare test loss versus training epochs on three types of data: (left) isotropic data with no feature correlation, (middle) data with a simple correlation structure, and (right) data with a complex correlation structure. The results show that while simple optimizers perform well in the absence of correlations, the proposed AdaGram methods (solid lines) excel as the data's correlation structure becomes more complex.}
    \label{fig:synth_epochs_results}
\end{figure}

The performance of our proposed methods on synthetic data with varying correlation structures is presented in Figure~\ref{fig:synth_epochs_results}. As expected, on isotropic data where feature correlations are absent, our methods are outperformed by simpler optimizers like AdaGrad and SGD. 
However, as correlations are introduced, the advantages of our approach become apparent.
On data with simple correlations, our AdaGram variants are highly competitive. 
In the complex correlations case, our methods demonstrate superior final performance, converging to a lower objective value than baselines.

\subsection{Evaluation on real datasets}
To explore AdaGram performance on real-world data, we conduct experiments on three UCI benchmark datasets: Australian, Heart, and Splice. 
The principal characteristics of these datasets are summarized in Table~\ref{tab:dataset_stats}. 
Our primary objective is to analyze convergence behavior with respect to wall-clock time, while also examining the time-to-optimum metric for each optimizer to achieve its best performance.

\begin{figure}[!h]
    \centering
    \includegraphics[width=\linewidth]{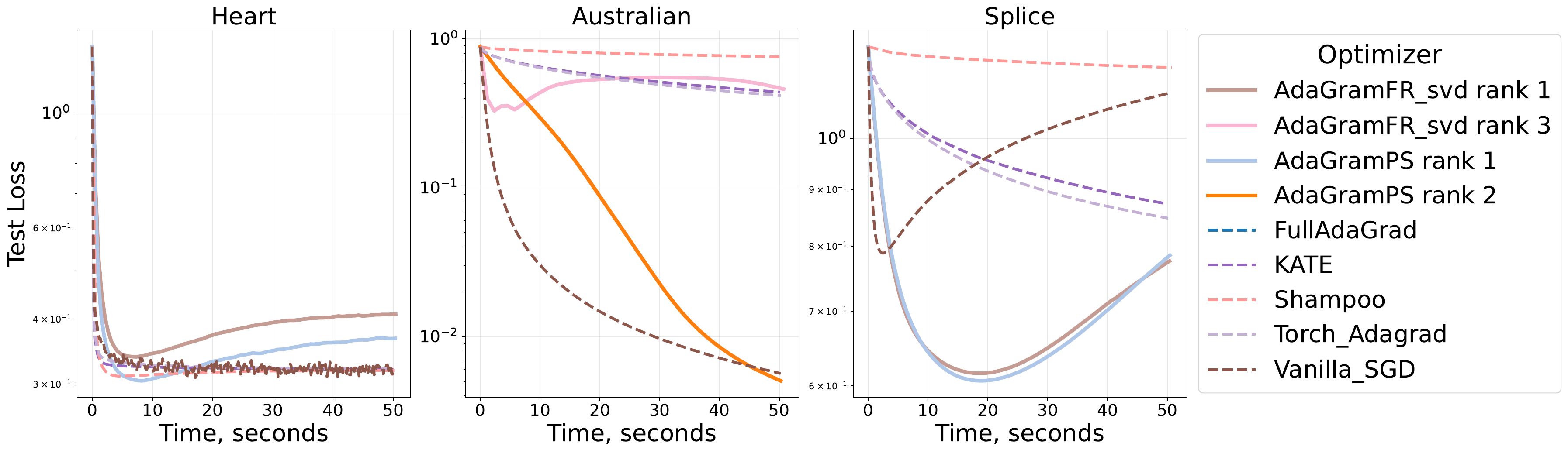}
    \caption{Comparison of optimization algorithms on the Heart, Australian, and Splice datasets. The proposed AdaGram variants (solid lines) converge to lower or comparable loss values compared to baseline methods.}
    \label{fig:loss_real_time}
\end{figure}

Figure~\ref{fig:loss_real_time} demonstrates the evolution of test loss as a function of wall-clock time, establishing that our SVD and projector splitting variants consistently achieve superior or equivalent performance compared to all baseline methods. 
Notably, grid search optimization reveals that the optimal rank for the AdaGram\_PS variant remains consistently low (rank 1 or 2 across all datasets). 
This low-rank property is theoretically significant, as it demonstrates that the method effectively captures essential gradient covariance structure while maintaining minimal computational complexity. 
The observed efficiency indicates favorable scalability characteristics for high-dimensional optimization scenarios.

\begin{figure}[!h]
    \centering
    \includegraphics[width=\linewidth]{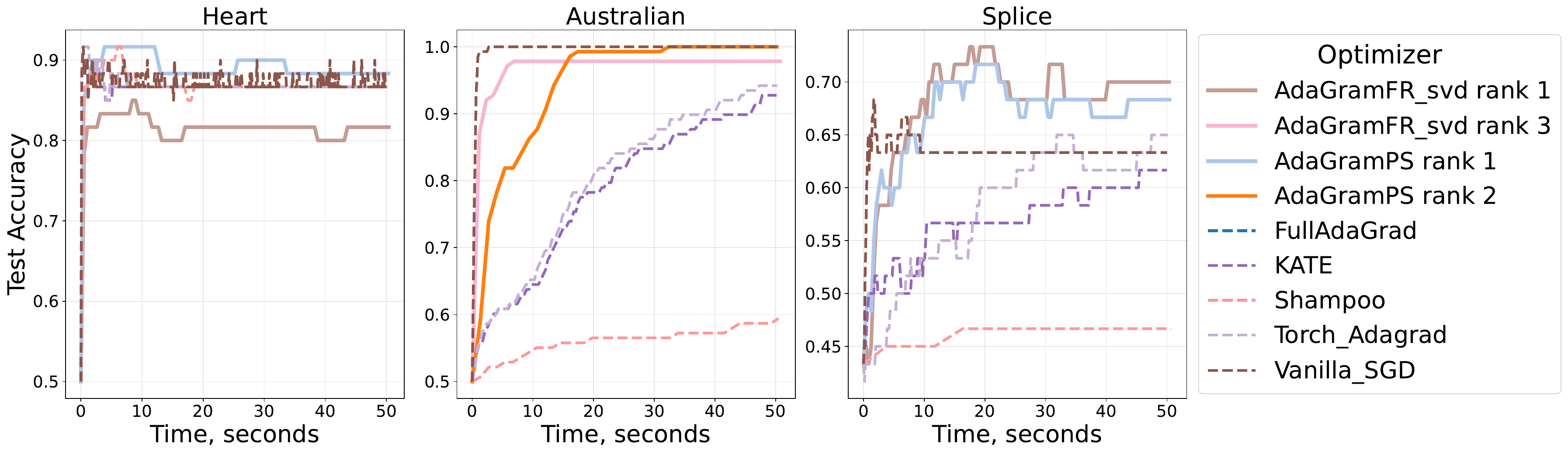}
    \caption{Performance of optimizers on the Heart, Australian, and Splice datasets. Our proposed methods (AdaGramFR and AdaGramPS) achieve higher or comparable final accuracy to competing methods, often with faster initial convergence.}
    \label{fig:acc_real_time}
\end{figure}

Plots, demonstrated in Figure~\ref{fig:acc_real_time}, indicate a clear advantage of our proposed methods. 
Across the datasets, they consistently reach a higher or equal final accuracy than the baseline optimizers. 
Crucially, they achieve this performance with no significant time overhead; in the case of the Splice dataset, convergence is notably faster.

\section{Conclusion}
In this work, we introduced AdaGram, a novel optimization algorithm that implements full-matrix adaptive preconditioning with high efficiency. 
By leveraging a fast symmetric factorization technique from~\cite{ambikasaran2014fast} and establishing a connection to matrix integrator theory, our method circumvents the prohibitive computational cost typically associated with full-preconditioner updates. 
Specifically, AdaGram avoids the need for costly matrix decomposition and inversion at each iteration, making it a practical approach for capturing gradient correlations.

Our empirical evaluation on Generalized Linear Models, particularly for a logistic regression task using a single-layer perceptron, demonstrates the efficiency and potential of AdaGram. 
The promising results validate our methodology and strongly motivate future work. 
Key directions include extending the AdaGram framework to more complex, arbitrary model architectures and investigating its performance on a broader range of machine learning tasks.

\bibliographystyle{unsrt}
\bibliography{lib}

\newpage
\appendix
\section{Appendix}

\subsection{Algorithms}
\label{algos}
\begin{algorithm}
\caption{AdaGrad, matrix case}
\label{alg:adagrad-matrix}
\begin{algorithmic}[1]
\REQUIRE $\eta > 0$, $\epsilon > 0$
\STATE Initialize $W_1 = \mathbf{0}_{m \times n}$; $G_0 = \epsilon I_{mn}$;
\FOR{$t = 1, \ldots, T$}
    \STATE Receive loss function $f_t : \mathbb{R}^{m \times n} \mapsto \mathbb{R}$
    \STATE Compute gradient $g_t = \text{vec}(\nabla f_t(W_t))$ \COMMENT{$g_t \in \mathbb{R}^{mn}$}
    \STATE Update preconditioner: $G_t = G_{t-1} + g_t g_t^\top$
    \STATE Reshape: $w_t = \text{vec}(W_t)$
    \STATE Update: $w_{t+1} = w_t - \eta G_t^{-1/2}g_t$
    \STATE Reshape: $W_{t+1} = \text{unvec}(w_{t+1})$
\ENDFOR
\end{algorithmic}
\end{algorithm}
\begin{algorithm}
\caption{Shampoo matrix case}
\label{alg:shampoo-matrix}
\begin{algorithmic}[1]
\REQUIRE $\eta > 0$, $\epsilon > 0$
\STATE Initialize $W_1 = \bm{0}_{m \times n}$; $L_0 = \epsilon \bm{I}_m$; $R_0 = \epsilon \bm{I}_n$
\FOR{$t = 1$ \TO $T$}
    \STATE Receive loss function $f_t : \mathbb{R}^{m \times n} \mapsto \mathbb{R}$
    \STATE Compute gradient $\bm{G}_t = \nabla f_t(W_t)$ \COMMENT{$\bm{G}_t \in \mathbb{R}^{m \times n}$}
    \STATE Update left preconditioner: $L_t = L_{t-1} + \bm{G}_t \bm{G}_t^\top$
    \STATE Update right preconditioner: $R_t = R_{t-1} + \bm{G}_t^\top \bm{G}_t$
    \STATE Compute update direction: $\Delta_t = L_t^{-1/4} \bm{G}_t R_t^{-1/4}$
    \STATE Update parameters: $W_{t+1} = W_t - \eta \Delta_t$
\ENDFOR
\end{algorithmic}
\end{algorithm}

\subsection{Proof of Theorem 1 (Recursive computation of a preconditioned gradient)}\label{theorem_proof}

\begin{equation}\label{eq:main}
    L_{t+1}^{-1}g = \frac{1}{\sqrt{1 + \|L_t^{-1}g\|^2}}\cdot L^{-1}_t g
\end{equation}

\begin{proof} Let us consider a base case $t=0$.
For $t=0$ we have:

\[
G_0 = \epsilon I \implies L_0 = \frac{1}{\sqrt{\epsilon}} I
\]

The identity immediately verifies the equality:
\begin{equation}
    L_1L_1^\top g_1 = G_1g_1 = \left(\epsilon I + g_1g_1^\top\right)g_1 = \left(\epsilon + \|g_1\|^2\right)g_1,
\end{equation}

That implies that:
\begin{equation}
L_1^{-1}g_1 =  \frac{1}{\sqrt{\left(\epsilon + \|g_1\|^2\right)}}g_1
\end{equation}

Let's prove the proposition for some time step $t>0$. By the means of Eq.~(\ref{eq:gram_matrix}) we have
\begin{equation}\label{eq:gprocess}
    G_{t+1} = G_t + g_{t+1} g_{t+1}^\top.
\end{equation}

Consider the following relation (indices are omitted for better comprehension):
\begin{equation}
    G + gg^\top = L_t\left(I + \bar g \bar g^\top \right)L_t^\top,
\end{equation}
where $\bar g = L_t^{-1}g$. We can further symmetrically factorize the middle term:
\[
\left(I + \bar g \bar g^\top \right) = PP^\top.
\]
By~\cite{ambikasaran2014fast}:
\begin{equation}
    P = I + \alpha\bar g \bar g^\top,
\end{equation}
where $\alpha$ is a constant satisfying the relation~\ref{eq:rootconst}
which leads to
\begin{equation}
    G + gg^\top = L_t\left(I + \alpha \bar g \bar g^\top \right)\left(I + \alpha \bar g \bar g^\top \right)^\top L_t^\top,
\end{equation}
or
\begin{equation}
   L_{t+1} = L_t\left(I + \alpha \bar g \bar g^\top \right).
\end{equation}
By applying the Sherman-Morrison-Woodbury identity, one gets
\begin{equation}
    L_{t+1}^{-1} = \left(I - \frac{\alpha}{1+\alpha\|\bar g \|^2} \bar g \bar g^\top \right)L_t^{-1}.
\end{equation}
From here,
\begin{equation}\label{eq:rootvec}
    L_{t+1}^{-1}g = \left(I - \frac{\alpha}{1+\alpha\|\bar g \|^2} \bar g \bar g^\top \right)\bar g = \left(1+\|\bar g\|^2 \right)^{-\frac{1}{2}}\bar g
\end{equation}
where we used (\ref{eq:rootconst}) to simplify expression within brackets. Expanding $\bar g$ in the last equation finally gives:
\begin{equation}\label{eq:main}
L_{t+1}^{-1}g = \frac{1}{\sqrt{1 + \|L_t^{-1}g\|^2}}\cdot L^{-1}_t g,
\end{equation}
which proves the relation (\ref{eq:matvec_recusrion}) for arbitrary timestep $t\geq0$.
\end{proof}

\subsection{Proof of Lemma (Recursive representation of the preconditioner inverse))}\label{lemma_proof}
\begin{proof}
We proceed by induction on $t$. 
\textbf{Base case:} For $t = 0$, we have $L_0^{-1} = (I - P_0 Q_0^T) L_0^{-1}$ with $P_0 = Q_0 = \emptyset$, which holds trivially.

\textbf{Inductive step:} Assume the formula holds for time $t$. From the Sherman-Morrison formula, we have:
\begin{align*}
L_{t+1}^{-1} &= L_t^{-1} - \frac{L_t^{-1} g_{t+1} g_{t+1}^T L_t^{-1}}{1 + g_{t+1}^T L_t^{-1} g_{t+1}}\\
&= L_t^{-1} - \beta_{t+1} \bar{g}_{t+1} \bar{g}_{t+1}^T L_t^{-1}\\
&= \left(I - \beta_{t+1} \bar{g}_{t+1} \bar{g}_{t+1}^T\right)L_t^{-1} 
\end{align*}

By the inductive hypothesis, $L_t^{-1} = (I - P_t Q_t^T) L_0^{-1}$. Substituting:
\begin{align*}
L_{t+1}^{-1} &= (I - P_t Q_t^T) L_0^{-1} \left(I - \beta_{t+1} \bar{g}_{t+1} \bar{g}_{t+1}^T\right)\\
&= (I - P_t Q_t^T - \beta_{t+1} \bar{g}_{t+1} \bar{g}_{t+1}^T + P_t Q_t^T \beta_{t+1} \bar{g}_{t+1} \bar{g}_{t+1}^T) L_0^{-1}
\end{align*}

Define $\tilde{g}_{t+1} = (I - Q_t P_t^T) \bar{g}_{t+1}$. Then:
\begin{align*}
P_t Q_t^T \bar{g}_{t+1} &= Q_t (P_t^T \bar{g}_{t+1})\\
&= P_t (Q_t^T \bar{g}_{t+1} - P_t^T \bar{g}_{t+1} + P_t^T \bar{g}_{t+1})\\
&= P_t (Q_t^T - P_t^T) \bar{g}_{t+1} + P_t P_t^T \bar{g}_{t+1}
\end{align*}

Using matrix distributive property and the definitions of $P_{t+1}$ and $Q_{t+1}$, we can verify that:

$$L_{t+1}^{-1} = \left(I - P_{t+1} Q_{t+1}^T\right) L_0^{-1},$$

where the updated matrices are:
$$P_{t+1} = [P_t \quad \beta_{t+1} \bar{g}_{t+1}], \quad Q_{t+1} = [Q_t \quad (I - Q_t P_t^T) \bar{g}_{t+1}]$$
\end{proof}

\end{document}